\documentclass{article}
\usepackage[utf8]{inputenc}
\usepackage{natbib}
\usepackage{graphicx}
\usepackage{booktabs}
\usepackage{multirow}
\usepackage{textcomp}
\usepackage[a4paper, total={6in, 8in}]{geometry}
\setcitestyle{authoryear, open={(},close={)}}

\title{ChemBoost: A chemical language based approach for protein - ligand binding affinity prediction}

\date{}
\author{
R{\i}za \"{O}z\c{c}elik$^{+, 1}$ \\ \texttt{riza.ozcelik@boun.edu.tr}
\and 
Hakime \"{O}zt\"{u}rk$^{+, 1}$ \\ \texttt{hakime.ozturk@boun.edu.tr} 
\and
Arzucan \"{O}zg\"{u}r$^{*, 1}$ \\ \texttt{arzucan.ozgur@boun.edu.tr} 
\and
Elif Ozkirimli$^{*, 2, 3}$ \\ \texttt{elif.ozkirimli@boun.edu.tr} 
}

\begin{document}

\maketitle
\noindent \textbf{1 Dept. of Computer Engineering, Bo\u{g}azi\c{c}i University, Istanbul, Turkey.\\
2 Dept. of Chemical Engineering, Bo\u{g}azi\c{c}i University, Istanbul, Turkey.\\
3 Data and Analytics Chapter, Pharma International Informatics, F. Hoffmann-La Roche AG, Switzerland} \\
$^+$\small{These authors contributed equally to the work.} \\
$^*$\small{To whom correspondence should be addressed.}

\section*{\centering{Abstract}}

Identification of high affinity drug-target interactions is a major research question in drug discovery. Proteins are generally represented by their structures or sequences. However, structures are available only for a small subset of biomolecules and sequence similarity is not always correlated with functional similarity. We propose ChemBoost, a chemical language based approach for affinity prediction using SMILES syntax. We hypothesize that SMILES is a codified language and ligands are documents composed of chemical words. These documents can be used to learn chemical word vectors that represent words in similar contexts with similar vectors. In ChemBoost, the ligands are represented via chemical word embeddings, while the proteins are represented through sequence-based features and/or chemical words of their ligands. Our aim is to process the patterns in SMILES as a language to predict protein-ligand affinity, even when we cannot infer the function from the sequence. We used eXtreme Gradient Boosting to predict protein-ligand affinities in KIBA and BindingDB data sets. ChemBoost was able to predict drug-target binding affinity as well as or better than state-of-the-art machine learning systems. When powered with ligand-centric representations, ChemBoost was more robust to the changes in protein sequence similarity and successfully captured the interactions between a protein and a ligand, even if the protein has low sequence similarity to the known targets of the ligand.

\textbf{Keywords:} \textit{chemical language, drug discovery, machine learning, proteins, virtual screening}

\section{Introduction}
Identification of high affinity drug-target interactions (DTI) powered by the available knowledgebase of protein - ligand interactions is an important first step in the drug discovery pipeline. Computational tools from structure-based drug design \citep{sledz2018protein} to quantitative structure–activity relationship (QSAR) \citep{bosc2019large} can accelerate this critical step by narrowing down potential binding partners. The prediction of binding affinity for novel interactions is still a challenging task because (i) representation of proteins and ligands in computational space is complicated by the inherent three-dimensional (3D) nature of the interaction \citep{limongelliligand}, (ii) as of April 2020, there are only around 17680 protein - ligand complex structures in PDBBind \citep{wang2004pdbbind}, (iii) the chemical space sampled by the currently available data (560K proteins in SwissProt \citep{apweiler2004uniprot}, 2M compounds in ChEMBL \citep{davies2015chembl}) is limited.


Similar to structure-based drug design studies \citep{barcellos2019pharmacophore, xue2019discovery}, machine learning methodologies can utilize the 3D structure information of  a protein - ligand complex to predict binding affinity \citep{gomes2017atomic, jimenez2018k, stepniewska2018development}. Such structure-based methodologies, however, are limited by the available structural information on the complex as stated in point (ii). Two dimensional (2D) graph convolutions can also be used to learn molecule representations \citep{jaeger2017mol2vec, mayr2018large, wu2018moleculenet}. However, graph-based methodologies rely on complex and challenging-to-interpret graph convolutional networks for representation learning \citep{gnne2019ying}. An attractive alternative is to use a string-based compound representation, which allows the application of the recent advances in natural language processing (NLP). The field of chemical linguistics that brings the chemistry and linguistics domains together has been growing since its inception in the 1960s \citep{garfield1961chemico}. Protein sequences have also been utilized to generate representations through different descriptors such as amino-acid composition, position-specific scoring matrix, amino-acid substitution matrices (e. g. BLOSUM, PAM) \citep{chen2018ifeature} as well as physicochemical properties of residues \citep{barley2018improved}, and eventually kmers \citep{asgari2015continuous} in NLP-inspired bio/cheminformatics applications. A recent review highlights the impact of NLP on drug discovery studies \citep{ozturk2020review}.

Simplified Molecular Input Line Entry System (SMILES) is a specialised syntax to represent molecules with an alphabet of over sixty characters. The SMILES representation can be used to reconstruct the 2D molecular graph, indicating its potential for encoding valuable molecular information. SMILES has been shown to perform as well as 2D representation-based graph convolutional embeddings in chemical property prediction \citep{jaeger2017mol2vec} and drug-target interaction prediction \citep{mayr2018large}. The SMILES representation has been successfully used in comparison or search algorithms \citep{cadeddu2014organic, wozniak2018linguistic}, as well as for different problems, including information retrieval  \citep{krallinger2017information} design of novel scaffolds to expand the chemical space, \citep{bostrom2018expanding} and prediction of chemical reaction outcomes \citep{schwaller2018found} and chemical properties \citep{convard1994smilogp}. Here, our aim is to treat the SMILES representation as a language and develop a machine learning and NLP based methodology for the task of  protein - ligand binding affinity prediction.

We propose ChemBoost, a novel chemical-language based approach that uses distributed ``chemical word" vectors for protein and/or ligand representation to predict interaction strength between targets and compounds. ChemBoost views SMILES strings as documents formed in a chemical language and processes the language units to create ligand and protein representations for affinity prediction. ChemBoost uses ``SMILESVec" \citep{ozturk2018novel}, a compound representation technique that utilizes the SMILES form. Distributed chemical word vectors are learned from a large corpus containing millions of SMILES strings. The chemical language-based nature of ChemBoost allows us to take advantage of the abundant textual data to learn chemical representations unlike structure-based learning algorithms \citep{gomes2017atomic, jimenez2018k, stepniewska2018development}, which are trained on a limited number of samples.

In the ChemBoost models, ligands are represented with their SMILESVec vectors. We investigate two different approaches for protein representation. The first approach is the standard approach for protein representation, where proteins are represented with their sequences. We utilize the Smith-Waterman and ProtVec \citep{asgari2015continuous} algorithms to obtain sequence-driven protein representations. The second approach is a ligand-centric approach, where proteins are represented with the distributed vectors of their ligands. 
Biologically and functionally similar proteins often bind to chemically similar ligands and ligand-centric protein similarity calculations  have been used in clustering and identifying similar proteins \citep{keiser2007relating, hert2008quantifying, keiser2009, ozturk2015classification, ozturk2018novel}. To the best of our knowledge, this is the first study that investigates the effectiveness of distributed chemical word vectors and ligand-centric protein representations for protein-ligand binding affinity prediction.
The effect of representing proteins through the chemical words of their known ligands or only high affinity ligands, as well as combining protein sequence based representation with ligand-centric representation are also examined. 

We compare the ChemBoost models with three state-of-the-art binding affinity prediction approaches: KronRLS \citep{pahikkala2014toward}, SimBoost \citep{he2017simboost} and DeepDTA \citep{ozturk2018deepdta}, all of which exploit the protein sequence information explicitly. SimBoost further integrates drug-target interaction network statistics to increase prediction accuracy. SimBoost and KronRLS employ traditional machine learning models for prediction, whereas DeepDTA is built upon a multi-layered CNN architecture. Thanks to its novel chemical language based and ligand-centric representations, ChemBoost achieves state-of-the-art performance using a simple predictor, eXtreme Gradient Boosting (XGBoost) \citep{chen2016xgboost}.

\section{Materials and Methods} \label{sec:methods}
\subsection{Data Sets}

To benchmark the performance of ChemBoost, we used the KIBA \citep{tang2014making} bioactivity data set of proteins from Kinase family and the BDB data set that we extracted from the BindingDB database, which contains proteins from different families. To compile the BDB data set, BindingDB was filtered based on the following criteria: (i) proteins and compounds with at least 6 and 3 interactions were kept, respectively, (ii) the experiment with high affinity was selected, if there were multiple instances of the same protein-ligand pair, and (iii) only the interactions with $K_d$ values were included, and then converted into $pK_d$ according to Equation  \ref{pkd}:
\begin{equation}\label{pkd}
    pK_d= -\log_{10} (\frac{K_d}{1e9})
\end{equation}
The BDB data set comprises about 31K interactions between 490 proteins and 924 compounds. The average number of ligands with known binding affinity values for a protein is 53.3, and the average number of ligands with strong binding affinity values (i.e., $pK_d$ value $>$ 7) for a protein is 7.3. 
The $pK_d$ threshold of strong/weak binding has been selected as in \citealp{he2017simboost}.

\begin{figure*}
    \centering
    \begin{tabular}{@{}c@{}}
        \includegraphics[width=0.5\linewidth]{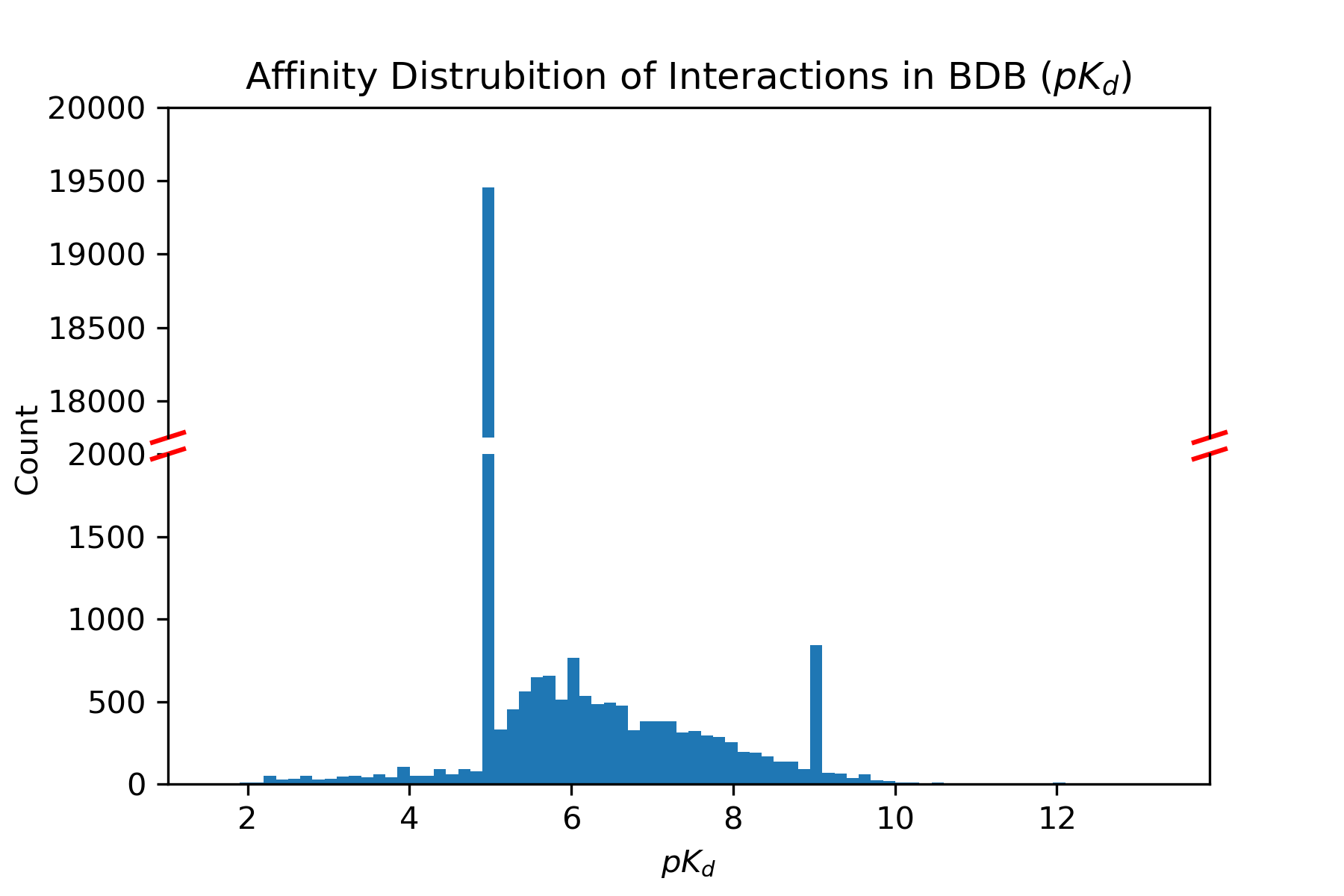} \includegraphics[width=.5\linewidth]{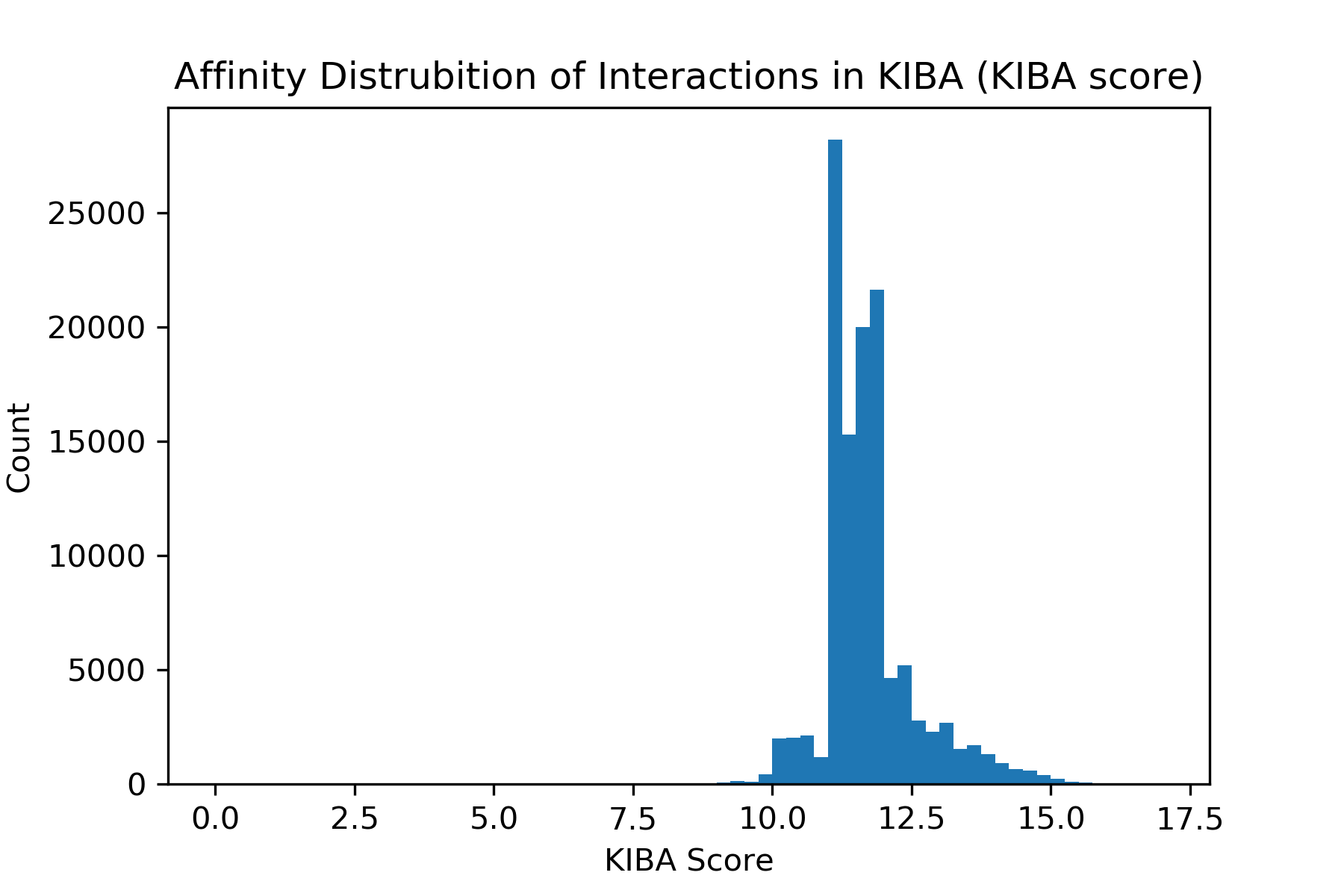}
    \end{tabular}
    \caption{Distribution of binding affinity values in BDB (left) and KIBA (right) data sets. The strong peak in BDB distributions is due to interactions reported as $pK_d <= 5$ and setting $pK_d$ to 5 for such weak interactions.}
    \label{fig:aff}
\end{figure*}

The KIBA data set comprises KIBA scores for about 118K interactions of 229 proteins and 2111 compounds as filtered by \cite{he2017simboost} such that all proteins and compounds have at least 10 interactions. KIBA score is a combination of different bioactivity measurement sources such as $K_i$, $K_d$ or $IC_{50}$ \citep{tang2014making}. The average number of ligands with known binding affinity values for a protein is 516.4, and the average number of ligands with strong binding affinity values (i.e., KIBA score > 12.1) for a protein is 99.2. Figure \ref{fig:aff} illustrates the distribution of the binding affinity values of the protein - ligand pairs in the BDB and KIBA data sets. We observe a strong peak at $pK_d = 5$ for BDB, since low affinities are frequently reported as $K_d >= 10000 $ $(pK_{d} <= 5)$. 

\subsection{Ligand representation} \label{sec:ligandrep}

We adopted SMILESVec \citep{ozturk2018novel} to represent ligands with distributed chemical word vectors based on their SMILES representations. Our underlying hypothesis is that, similarly to natural languages where documents and sentences are composed of words, SMILES strings constitute a domain-specific language composed of chemical words. SMILESVec utilizes Word2vec \citep{mikolov2013distributed} to learn distributed chemical word vectors from a large SMILES corpus. 

Word2Vec assumes that words that appear frequently in  similar contexts have higher semantic similarity, where context is defined as a set of words within a window frame. By training a single layered neural network to predict either the target word given the context (the CBOW model) or the context given a target word (the Skip-Gram model), Word2Vec learns a vector for each word in the training corpus. These distributed vectors of words that occur in similar contexts are close to each other in the Euclidean space. SMILESVec treats SMILES strings of compounds as sentences and 8-mers as words to learn distributed chemical word vectors.

The SMILESVec of a compound is described in Equation \ref{eq:1}, in which $n$ is equal to the number of chemical words ($cw$) extracted from the SMILES string of a compound and $vector(cw_k)$ represents the 100-dimensional (100D) embedding of the $k^{th}$ chemical word \citep{ozturk2018novel}. The compound is then described as the average of the vectors of the chemical words in its SMILES representation. The averaging method is also adopted by other recent studies that learn embeddings from biological and chemical data \citep{asgari2015continuous,jaeger2017mol2vec}. 

\begin{equation} \label{eq:1}
    SMILESVec = vector(ligand) =  \frac{\sum_{k=1}^n vector(cw_k)}{n} 
\end{equation}

One of the most important steps in SMILESVec ligand representation is the identification of the chemical words. In this study, besides the previously used k-mer approach for word identification, we applied and evaluated the Byte Pair Encoding (BPE) approach, which has been originally proposed for word boundary detection in the NLP field \citep{sennrich2015neural}. Table \ref{tab:ampicillinwords} provides examples of the words that are extracted from the SMILES of 7-Chloro-1-(3,4-dichlorophenyl)-6-methoxy-3,4-dihydroisoquinoline, ``COc1cc2CCN=C(c3ccc(Cl)c(Cl)c3)c2cc1Cl" with both word identification techniques, where k is set to 8 for k-mers. 

\begin{table}
\caption{Example words extracted from the SMILES of \textit{7-Chloro-1-(3,4-dichlorophenyl)-6-methoxy-3,4-dihydroisoquinolin} (COc1cc2CCN=C(c3ccc(Cl)c(Cl)c3)c2cc1Cl) using 8-mers and BPE.}
\centering
        \begin{tabular}{@{}ll@{}}
            \toprule
            \textbf{Method} & \textbf{Words} \\ \midrule
            $k$-mer (i.e. 8-mer) & \begin{tabular}[c]{@{}l@{}} COc1cc2C, Oc1cc2CC, c1cc2CCN, ..., 3)c2cc1C, \\  )c2cc1Cl 
        \end{tabular} \\ \midrule
            BPE   & \begin{tabular}[c]{@{}l@{}}  COc1cc2, CCN=C(, c3ccc(Cl)c(Cl)c3), \\
            c2cc1Cl \end{tabular}  \\ \bottomrule
        \end{tabular}
 \label{tab:ampicillinwords}
\end{table}

\subsubsection{$k$-mers}
To extract chemical words as k-mers, we used a sliding window approach as in our previous work \citep{ozturk2018novel}. We traversed a window of length $k$ over the SMILES string of a ligand and extracted all overlapping SMILES substrings of length $k$. We represented each ligand with the resulting \textit{k-mers} in which each \textit{k-mer} is a chemical word. We set $k$ to $8$, since 8-mers have been shown to outperform other options ranging from 4-mers to 12-mers \citep{ozturk2018novel}. We then employed the Word2Vec algorithm to learn embeddings for these chemical words by training on approximately 1.7M canonical SMILES strings that were collected from the ChEMBL database (vChEMBL23). We used the \texttt{gensim} implementation \citep{rehurek2011gensim} of Word2Vec with the \texttt{Skip-Gram} approach and the size of the vectors was set to the default value of 100. 

\subsubsection{Byte Pair Encoding (BPE)}
Byte Pair Encoding (BPE) is a compression technique \citep{gage1994new} that was adopted to the word segmentation task in the NLP field for extracting the words or tokens of a language given a large corpus \citep{sennrich2015neural}. The BPE algorithm is based on the assumption that frequently occurring substrings are meaningful and can be treated as language units. To identify the most frequent substrings, BPE starts with an initial vocabulary of characters (unigrams) in the corpus and counts the frequency of bigrams. The most frequent bigram is included to the vocabulary as a token and bigrams are counted again while treating the bigrams in the vocabulary as a single character. The iterations continue until a pre-defined iteration limit or vocabulary size is reached. The resulting vocabulary constitutes the words extracted from the training corpus. A recent study successfully used ``bio-words" extracted from a corpus of protein sequences via the BPE algorithm  and showed its effectiveness for the task of protein-protein interaction prediction \citep{wang2019high}. 

In this work, we used BPE to learn the chemical words from a corpus of SMILES strings. The BPE algorithm was trained on the vChEMBL23 SMILES corpus with the vocabulary size of 20K, character coverage of 0.99 and maximum word length of 100 characters. The \texttt{sentencepiece} library in Python (\texttt{github.com/google/sentencepiece/}) was utilized. In order not to omit the symbols in the rich vocabulary of SMILES strings, \texttt{number\_split'} and \texttt{unicode\_split} parameters were set to \texttt{False}.  

\subsubsection{Benchmark Representations}
Besides the chemical-language based vectors, we used molecular access system (MACCS) \citep{durant2002reoptimization} keys and Morgan fingerprints \citep{morgan1965generation} to represent the ligands. MACCS vectors are lists of chemical binary properties about the molecule, whereas Morgan fingerprints are obtained through encoding 2D structure information with an iterative process. We used rdkit \citep{landrum2006rdkit} to compute both representation type and used 166-dimensional MACCS vectors and 2048-dimensional Morgan fingerprints with radius 2.

\subsection{Protein representation}

We represented proteins in a ligand-centric way, in which the average of the word embeddings of the chemical words in their interacting ligands \citep{ozturk2018novel} were used to construct the protein vector. The chemical words that were used to build the SMILESVec representation were either 8-mers or BPE-based words. Consequently, representation of proteins changed according to the word identification technique that was used to describe SMILESVec. 

We represented proteins either using all ligands with a reported affinity, similarly to our previous work \citep{ozturk2018novel}, or only by their strong affinity ligands. In the high affinity approach, we considered the chemical words of all ligands with a reported affinity score, if a protein with no high affinity ligands is encountered. For BDB, the $pK_d$ value of 7 was selected as the threshold to divide the ligands into strong-binding and weak-binding classes ($pK_d > 7$ strong binding), whereas for the KIBA dataset, KIBA score of 12.1 was set as the threshold as in \citep{he2017simboost}. In addition, we filtered whole BindingDB to create a database of high affinity protein-ligand pairs including experiments with  $K_i$, $IC_{50}$ and $EC_{50}$ measurements. High affinity threshold for each measurement was chosen based on the distribution of affinity values and the pairs that are already in BDB or KIBA were removed. The resulting high affinity protein-ligand database was utilized to enrich the ligand-centric protein representations and reduce the number of proteins with no high affinity ligands in the benchmark BDB and KIBA data sets.

In addition to the proposed ligand-centric protein representation framework, we represented proteins with two existing methods: Smith-Waterman (SW) and ProtVec \citep{asgari2015continuous}. SW represents each protein with a vector that comprises its normalized SW similarity score with every other protein in the data set. ProtVec, on the other hand, is a Word2Vec based approach that represents proteins with 100-dimensional embeddings trained on 3-residue subsequences of a large protein sequence corpus.  We used representations provided
by the study. \citep{asgari2015continuous}

\subsection{eXtreme Gradient Boosting (XGBoost)}
Gradient boosting tree (or gradient boosting machine) is a popular machine learning algorithm that is an ensemble of sequential trees in which a given tree \textit{t} aims to learn from the misclassified samples of the previous $t-1$ trees by assigning them higher weights \citep{friedman2001greedy}. eXtreme Gradient Boosting (XGBoost), proposed by \cite{chen2016xgboost}, is built on the gradient boosting tree algorithm and is a regularized and scalable version of the original algorithm to avoid over-fitting. XGBoost has became a popular choice even surpassing popular deep neural networks as the statistics from Kaggle challenges indicate \citep{chen2016xgboost}. XGBoost has also recently been employed in different bioinformatics tasks such as QSAR studies \citep{sheridan2016extreme} and prediction of physical chemistry properties \citep{wu2018moleculenet}. 

A tree ensemble in XGBoost can be formulated as follows:

\begin{equation} \label{eq:boost}
    \hat{y}_i =  \sum_{k=1}^{K} f_k(x_i),  f_k \in F
\end{equation}
where a protein-ligand pair  $i$ is represented with a vector of ${x_i}$ and $K$ is the number of trees in the ensemble, whereas $F$ is the set of all possible trees to predict the respective affinity value $\hat{y}_i$ \citep{chen2016xgboost}.

\subsection{State-of-the-art models}
We compared the methods presented here with three state-of-the-art methodologies. We adopted the Kronecker-Regularized Least Squares (KronRLS) algorithm that predicts binding affinity while representing both proteins and ligands with their pairwise similarity score matrices \citep{pahikkala2014toward}. In order to compute the similarity between proteins and between compounds, the Smith-Waterman (S-W) algorithm and the PubChem structure clustering tool \texttt{(http://pubchem.ncbi.nlm.nih.gov)} were utilized, respectively. Second, we employed SimBoost, which is a gradient boosting machine based method \citep{he2017simboost} that depends on feature engineering of ligands and proteins utilizing information such as similarity and network-inferred statistics. Last, we compared our results with DeepDTA, which is a multi-layered Convolutional Neural Network (CNN) \citep{ozturk2018deepdta} based prediction model. The SMILES representations of ligands and  sequences of proteins are provided as inputs to DeepDTA to predict the binding affinity. 
All of these baselines utilize protein sequence features explicitly.

\subsection{Evaluation}
As we formulated protein-ligand binding affinity prediction as a regression problem, the performances of the presented models were measured by calculating the Concordance Index (CI), Mean Squared Error (MSE), Root Mean Squared Error (RMSE), and R$^2$ metrics metrics. CI is described in Equation \ref{eq:10} as follows \citep{gonen2005concordance}:
\begin{equation} \label{eq:10}
    CI = \frac {1} {Z} \sum_{\delta_x > \delta_y} h (b_x - b_y)
\end{equation}
\noindent where $b_x$ is the prediction value for the larger affinity $\delta_x$, $b_y$ is the prediction value for the smaller affinity $\delta_y$, $Z$ is a normalization constant, $h(m)$ is the step function, which is equal to 0 if $m <0$, 0.5 if $m = 0$, and 1 if $m > 0$  \citep{pahikkala2014toward}

MSE measures the average squared difference between the predicted values ($p$) and the actual values ($y$) where $n$ is the number of samples (Equation \ref{eq:11}).
\begin{equation} \label{eq:11}
    MSE= \frac{1}{n}\sum_{k=1}^{n}(p_k - y_k)^2   
\end{equation} 

We also computed RMSE by taking the square root of Equation \ref{eq:11} in order to reduce MSE to the order of the actual affinity scores.

R$^2$ is a scale-invariant prediction quality metric, unlike MSE and RMSE, that measures how much of the variance in the actual values is explained by the predictions. We computed R$^2$ using the following equation:

\begin{equation} \label{eq:r2}
    R^{2}=1-\frac{\sum_{k=1}^{n}\left(y_{k}-p_{k}\right)^{2}}{\sum_{k=1}^{n}\left(y_{k}-\bar{y}\right)^{2}}
\end{equation} 
\noindent where $\bar{y}$ is the mean of the actual values.
\subsection{Experimental Settings}
We used the same training and test folds that were used in our previous work \citep{ozturk2018deepdta}, but used ChEMBL canonical SMILES in ligand representation to comply with SMILESVec vectors. In these folds, both data sets were randomly divided into six equal parts and one part was separated as the independent test set. The remaining folds were used to determine the model hyper-parameters, such as learning rate and number of trees via five-fold cross validation. The hyper-parameter combination with which we obtained the best MSE value based on the cross-validation results over the training set was selected to model the test set. To report the performance of the models on the test set, we first trained the models on 4 folds of the training set 5 times by masking a different fold at each time. Then, we evaluated each of the 5 models on the test set and reported their average performance values alongside the standard deviation.

\section{Results and Discussion} \label{sec:results}

With this study, we introduce ChemBoost, a novel protein-ligand binding affinity prediction approach in which bio-molecules are represented through chemical-language based distributed vectors (SMILESVecs). MACCS keys and Morgan fingerprints are used to compare the text-based SMILESVec to widely adopted fingerprints in ligand representation. We investigate ligand-centric protein representations for affinity prediction, by averaging the distributed chemical word vectors of the ligands of a protein. We further compare the impact of three approaches in representing proteins: (i) using all ligands with a reported affinity, (ii) using only strong binding (SB) ligands and (iii) utilizing an additional, non-redundant high-affinity protein-ligand interaction database. SW and ProtVec are used as alternative protein representations to assess the effectiveness of ligand-centric protein representation in comparison with the use of amino-acid sequences. In addition, the effect of the chemical-word segmentation approach is examined by comparing 8-mers \cite{ozturk2018novel} with Byte Pair Encoding (BPE) based segments \citep{sennrich2015neural}. We evaluate different predictive models on BDB and KIBA using CI and MSE as evaluation metrics and compare ChemBoost-based models with three state-of-the-art affinity prediction models, two traditional machine learning based systems, namely KronRLS and SimBoost, and a deep-learning based approach, DeepDTA. We compute CI, MSE, RMSE, and R$^2$ scores, but compare the models based on only CI and MSE for readability, because MSE is highly parallel with RMSE and R$^2$. We report the significance levels of comparisons using paired t-test for models with close results.

To show that representing proteins with their high affinity ligands is effective at capturing functional similarities, which in turn may be useful for affinity prediction, we clustered the proteins in BDB and KIBA using SMILESVec-based ligand centric representations. 
The clusters are obtained via hierarchical clustering with ward linkage \citep{jones2001scipy}. The resulting clusters contain proteins from same or similar families and often share functional similarities. For instance, in the KIBA data set, more than half of Cluster 1 are Kinase C family (EC=2.7.11.13) proteins, grouped together because of their high affinity interactions with the same or similar ligands.
In Cluster 8 of the BDB data set, 15 out of 22 proteins belong to the rhodopsin family of GPCRs, four of them are chromo-domain, two of them are Rho-associated protein kinase and Ribosomal protein S6 kinase beta-1. 
The  ligands of each protein in this cluster contained the sub-sequence '@H](NC(=O)[C', which appears in only one tenth of the compounds in the BDB data set.
This information indicates that this ''chemical word" is a common signature of the proteins in this cluster, and thus SMILESVec-based protein representation was able to connect proteins from the same family through such chemical words. 

\subsection{Investigation of chemical-language based protein and ligand representations}

\begin{table*}
\caption{CI, MSE, RMSE and R$^2$ scores of ChemBoost models on BDB (top) and KIBA (bottom). Each model is trained with 5 different training sets and test set performance is measured for each trained model. Mean test set performance values and the standard deviations (in parenthesis) are reported} 
\begin{center}
\resizebox{\textwidth}{!}{

\begin{tabular}{llllcccc}

\toprule
\multicolumn{3}{c}{\textbf{Model}} & \multicolumn{4}{c}{\textbf{Scores}} \\ \midrule

&\textbf{Name} & \textbf{Protein Representation} & \textbf{Ligand Representation} &  \textbf{CI} & \textbf{MSE} & \textbf{RMSE} & \textbf{R$^2$} \\ \midrule

\parbox[t]{.2mm}{\multirow{15}{*}{\rotatebox[origin=c]{90}{\textbf{BDB}}}} & Model (S1) & SW & - & 0.687 (0.002) & 1.037 (0.006)& 1.018 (0.003)& 0.265 (0.004) \\

&Model (S2) & - & SMILESVec (8mer) & 0.773 (0.002) & 0.876 (0.005) & 0.936 (0.002) & 0.379 (0.003) \\

&Model (R1) & SW  &  Random  & 0.859 (0.002)  & 0.512 (0.005)  & 0.716 (0.003)  & 0.637 (0.003) \\

&Model (R2)  & Random  & SMILESVec (8mer) & 0.849 (0.002)  & 0.537 (0.009)  & 0.733 (0.006) & 0.619 (0.006) \\

&Model (F1) & SW & MACCS & 0.811 (0.003) & 0.817 (0.016) & 0.904 (0.009) & 0.421 (0.012) \\

&Model (F2) & SW & Morgan & 0.819 (0.002) & 0.767 (0.016) & 0.874 (0.009) &  0.458 (0.011) \\

&Model (1) & SW & SMILESVec (8mer) & 0.873 (0.001)  & 0.439 (0.008)  & 0.662 (0.006) & 0.689 (0.006) \\

&Model (2) & ProtVec & SMILESVec (8mer) & 0.854 (0.002) & 0.512 (0.004) & 0.716 (0.003) & 0.637 (0.003) \\

&Model (3) & ProtVec & SMILESVec (BPE) & 0.849 (0.002) & 0.548 (0.008) & 0.740 (0.006) & 0.611 (0.006) \\

&Model (4) & SMILESVec (all, 8mer) & SMILESVec (8mer) & 0.847 (0.001) & 0.524 (0.006) & 0.724 (0.004) & 0.628 (0.004) \\

&Model (5) & SMILESVec (SB, 8mer) & SMILESVec (8mer) & 0.845 (0.002) & 0.478 (0.005) & 0.692 (0.004) & 0.661 (0.003) \\

&Model (6) & SMILESVec (SB, BPE) & SMILESVec (BPE) & 0.842 (0.001) & 0.497 (0.007) & 0.705 (0.005) & 0.647 (0.005) \\

&Model (7) & SMILESVec (BindingDB, SB, 8mer)  & SMILESVec (8mer) & 0.856 (0.001) & 0.454 (0.007) & 0.674 (0.006) & 0.678 (0.005) \\

&Model (8) & SW \& SMILESVec (SB, 8mer) & SMILESVec (8mer) & 0.873 (0.001) & 0.420 (0.004) & 0.648 (0.003) & 0.702 (0.003) \\

&Model (9) & SW \& SMILESVec (BindingDB, SB, 8mer) & SMILESVec (8mer) & 0.871 (0.002) & 0.420 (0.007) & 0.648 (0.005) & 0.702 (0.005) \\ \midrule

\parbox[t]{2mm}{\multirow{15}{*}{\rotatebox[origin=c]{90}{\textbf{KIBA}}}} & Model (S1) & SW & - & 0.683 (0.000) & 0.585 (0.000) & 0.765 (0.000) & 0.139 (0.001) \\

& Model (S2) & - & SMILESVec (8mer) & 0.699 (0.000) & 0.425 (0.001) & 0.652 (0.000) & 0.374 (0.001) \\

&Model (R1) & SW & Random & 0.803 (0.001) & 0.276 (0.002) & 0.525 (0.002) & 0.594 (0.003) \\

&Model (R2) & Random & SMILESVec (8mer) & 0.815 (0.001) & 0.258 (0.002) & 0.508 (0.002) & 0.621 (0.002) \\

&Model (F1) &  SW &  MACCS &  0.829 (0.001) &  0.222 (0.002) & 
0.471 (0.002) & 0.674 (0.002) \\

&Model (F2) &  SW &  Morgan &  0.847 (0.001) &  0.186 (0.002) &  0.431 (0.002) & 0.727 (0.003) \\

&Model (1) & SW & SMILESVec (8mer) & 0.837 (0.001) & 0.203 (0.002) & 0.450 (0.002) & 0.702 (0.003) \\

&Model (2) & ProtVec & SMILESVec (8mer) & 0.818 (0.001) & 0.244 (0.001) & 0.494 (0.001) & 0.641 (0.001) \\

&Model (3) & ProtVec & SMILESVec (BPE) & 0.814 (0.001) & 0.252 (0.002) & 0.502 (0.002) & 0.630 (0.003) \\

&Model (4) & SMILESVec (all, 8mer) & SMILESVec (8mer) & 0.823 (0.001) & 0.243 (0.003) & 0.493 (0.003) & 0.642 (0.004) \\

&Model (5) & SMILESVec (SB, 8mer) & SMILESVec (8mer) & 0.829 (0.001) & 0.221 (0.001) & 0.470 (0.001) & 0.675 (0.001) \\

&Model (6) & SMILESVec (SB, BPE) & SMILESVec (BPE) & 0.825 (0.001) & 0.227 (0.001) & 0.477 (0.002) & 0.665 (0.002) \\

&Model (7) & SMILESVec (BindingDB, SB, 8mer) & SMILESVec (8mer) & 0.829 (0.001) & 0.223 (0.001) & 0.472 (0.001) & 0.672 (0.001) \\

&Model (8) & SW \& SMILESVec (SB, 8mer) & SMILESVec (8mer) & 0.837 (0.001) & 0.206 (0.001) & 0.454 (0.002) & 0.697 (0.002) \\

&Model (9) & SW \& SMILESVec (BindingDB, SB, 8mer) & SMILESVec (8mer) & 0.836 (0.001) & 0.207 (0.002) & 0.455 (0.002) & 0.696 (0.002) \\ \bottomrule

\end{tabular}
}
\end{center}

\label{tab:bdb_kiba_merged_res}
\end{table*}

In this subsection, we inspected the effectiveness of SMILESVec-based ligand and protein representations in the affinity prediction problem. We trained each model five times with different folds of the training set. We measured the performance on the test set for each trained model and reported the average results in terms of CI and MSE. We report the experiment results on the BDB and KIBA (Table \ref{tab:bdb_kiba_merged_res}) data sets utilizing XGBoost as the prediction algorithm for all cases.

\paragraph{Ligand representation} We first tested the impact of ligand representation by creating two baselines. Model (S1) does not utilize any ligand information, whereas Model (R1) represents each ligand with a 100-dimensional random vector sampled from a uniform distribution over [0, 1). In both models, proteins are represented with SW vectors. Model (R1) outperformed Model (S1) on both data sets, with respect to both evaluation metrics, emphasizing the requirement for ligand information to build a successful predictive model. As each random vector is associated with a particular ligand, their success in prediction performance can be linked to ligand-specificity. We suggest that these random vectors encode the identity of the ligands and thus, improve model performance, even though the content of the representation vector is pure noise.

We then assessed the performance of MACCS keys by comparing Model (R1) and Model (F1). Both models represent proteins with SW vectors, whereas Model (R1) uses random vectors for ligand representation and Model (F1) uses MACCS keys. Model (F1) outperformed Model (R1) on KIBA in terms of both MSE and CI, whereas on BDB, Model (R1) achieved better scores. We then evaluated the effectiveness of Morgan fingerprints by comparing Model (F1) and Model (F2) and observed that Morgan fingerprints are superior to MACCS keys for affinity prediction, since they yielded higher scores on both datasets. However, similar to MACCS keys, the performance of Morgan fingerprints is lower than random vectors on BDB, suggesting that fingerprint-based representations are not sufficiently distinctive for the ligands in BDB.

In order to analyze the performance of distributed chemical word vectors based approach in ligand representation, we compared Model (1) and Model (F2). While both models represent proteins using SW, Model (1) represents ligands with SMILESVec and Model (F2) represents ligands with Morgan fingerprints. Morgan vectors yielded the best performance among all models on KIBA in terms of both metrics, whereas their performance is lower than random vectors on BDB. SMILESVecs, on the other hand, achieved high performance on both datasets, suggesting that they are more consistent representations for the binding affinity prediction task.

We designed Models (2), (3), (5) and (6) to investigate the effect of different chemical-word tokenization techniques on prediction performance. Models (2) and (3) adopted ProtVec for protein representation, whereas Models (5) and (6) used the average of chemical-word vectors of proteins with high-affinity ligands. Comparing Model (2) with (3) and Model (5) with (6), we observed that 8-mer based representations achieved higher performance than their BPE counterparts, indicating that 8-mers might be more suitable language units for the affinity prediction task. Consequently, we decided to utilize 8-mers as chemical words in the remaining experiments.

\paragraph{Protein representation} Here, we also designed two models to show the impact of protein representation on the affinity prediction problem. We constructed Model (S2) which does not utilize any protein-related information and Model (R2) that represents each protein with a random vector that is sampled from a uniform distribution over [0, 1). Both models represented the ligands with SMILESVecs. Model (R2) outperformed Model (S2) with respect to both MSE and CI, on both data sets. Similar to our experiment with ligand representation, we suggest that random vectors are interpreted as unique fingerprints for each protein by the prediction algorithm. This information boosted the performance of the system, validating the necessity of protein information for affinity prediction modeling.

To test the effectiveness of different protein representation techniques, we first compared protein-specific random vectors (Model (R2)) with SW (Model (1)) in which both models described ligands with SMILESVecs. The use of SW representation improved prediction performance on both data sets for both metrics, demonstrating the advantage of SW vectors over random vectors. Then we compared SW (Model (1)) with ProtVec (Model (2)) in which SW outperformed ProtVec on both data sets for both evaluation metrics. This result shows that SW, which includes sequence similarity and amino acid physicochemical information, is a better alternative for representing proteins in affinity prediction, when combined with SMILESVec.

Although each binding affinity measurement provides a valuable data point for learning algorithms, most often, the mechanism of bimolecular interaction is accurately described by high affinity interactions \cite{martin1998protein}. Therefore, to verify that the contribution of high affinity ligands is more informative for ligand-centric protein representation, we compared the use of chemical words of all ligands with a reported affinity value (Model (4)) to the use of chemical words of high affinity ligands (Model (5)). In our experiments, all known ligands of a protein are the ones for which the affinity value to the protein is reported. In both cases, 8-mer based SMILESVecs were used to represent the ligands. The results illustrated that using high affinity ligands in protein representation outperformed the model in which all known ligands are utilized in terms of all metrics on both data sets.
These results emphasized that considering strong binding ligands in protein representation improves the prediction performance, motivating us to construct Model (7).

In Model (7), the number of high-affinity ligands incorporated in protein representation is increased by including high affinity protein-ligand pairs obtained from different experimental measurements in BindingDB (e.g. $K_i$, $IC_{50}$ etc.). These pairs are used in the protein representation in addition to the ones that are already in the training set. We
compared Model (7) with Model (5) using a paired t-test at 99\% significance and observed significant improvement in MSE and CI on the BDB dataset, whereas the same test indicated that their performance are on par with each other on KIBA. The improvement for the BDB may be due to the higher increase in the number of high-affinity ligands of a protein. The number of strong binding ligands of a protein increased 17.1 times in BDB but only 2.3 times in KIBA with the inclusion of the additional data.

On the other hand, Model (1), which utilizes SW in protein representation, outperformed Model (7) with respect to both metrics for both data sets, suggesting the merit of SW over ligand-centric protein representation. Then, we constructed Model (8) and Model (9) where we concatenated SW vectors with ligand-centric representation to incorporate both amino-acid sequence and ligand binding information. A paired t-test with 99\% confidence did not  indicate a distinction between Models (8) and (9), despite Model (9) utilizing an external database of high-affinity protein-ligand pairs. 

Model (8), however, provided an improvement over Model (6) and Model (9) outperformed Model (7), emphasizing that integration of SW brings in complementary information to ligand-centric representation on both data sets. Last, we compared the hybrid protein representation (Model (9)) with SW (Model (1)).Though the models performed similarly in terms of CI, MSE indicated a distinction between two representations. On the BDB data set, concatenating SW and ligand-centric vectors (Model (9)) yielded a better MSE than Model (1) with 95\% confidence, whereas on KIBA, the concatenation resulted in a performance decrease. This can be due to higher physicochemical similarity of Kinases in KIBA, compared to the heterogeneous structure of BDB.

We conclude that both protein and ligand information is  indispensable for predicting binding affinities of protein - ligand pairs, and distributed chemical word vectors (SMILESVecs) can be successfully utilized in representing these biomolecular entities.

Distributed vectors for ligand representation combined with SW or ligand-centric protein vectors capture the necessary information for the targets. The results indicate that the combination of strong affinity ligands-based protein representation with SW can improve the predictive performance for data sets with high protein diversity (i.e. proteins from different families). On the other hand, SW vectors might be more suitable for protein representation in data sets with low protein diversity (i.e. proteins from the same family). We also suggest that 8-mers can be used as chemical words in affinity prediction, since it is a simple and effective technique that performs better than the more complicated BPE approach.

\subsection{Comparison of ChemBoost and State-of-the-Art Models}
\label{sec:sota}
\begin{table}
\caption{CI and MSE scores of the state of the art affinity prediction models and ChemBoost on BDB (top) and KIBA (bottom). Here ChemBoost refers to the model in which the SMILESVec of a protein is obtained through the SMILES representations of its high affinity ligands and SW scores. Results are computed similarly to Table \ref{tab:bdb_kiba_merged_res}.} 
\centering
\begin{tabular}{llcccc}
\toprule

& \textbf{Model} &  \textbf{CI} & \textbf{MSE} & \textbf{RMSE} & \textbf{R$^2$}\\ \midrule

\parbox[t]{2mm}{\multirow{4}{*}{\rotatebox[origin=c]{90}{\textbf{BDB}}}} & KronRLS & 0.815 (0.003) & 0.939 (0.005) & 0.969 (0.002) & 0.334 (0.003) \\

& SimBoost & 0.855 (0.004) & 0.501 (0.026) & 0.707 (0.018) & 0.645 (0.018) \\

& DeepDTA & 0.863 (0.007) & 0.397 (0.011) & 0.630 (0.009) & 0.719 (0.008) \\

& ChemBoost & 0.871 (0.002) & 0.420 (0.007) & 0.648 (0.005) & 0.702 (0.005) \\ \midrule

\parbox[t]{2mm}{\multirow{4}{*}{\rotatebox[origin=c]{90}{\textbf{KIBA}}}} & KronRLS & 0.785 (0.001) & 0.411 (0.001) & 0.641 (0.001) & 0.395 (0.002) \\

& SimBoost & 0.836 (0.001) & 0.224 (0.001) & 0.473 (0.001) & 0.671 (0.001) \\

& DeepDTA & 0.846 (0.002) & 0.215 (0.005) & 0.464 (0.006) & 0.683 (0.008) \\

& ChemBoost & 0.836 (0.001) & 0.207 (0.002) & 0.455 (0.002) & 0.696 (0.002) \\

\bottomrule 

\end{tabular}
\label{tab:sota}
\end{table}

We compared one of the best ChemBoost models (Model (9)) with three state-of-the-art drug-target affinity prediction models: (i) KronRLS \citep{pahikkala2014toward}, a regularized linear regression model that uses SW similarity of protein sequences and PubChem structure similarity of ligands, (ii) SimBoost \citep{he2017simboost}, a gradient boosting tree based prediction algorithm that utilizes network statistics of protein-ligand interactions, in addition to the similarity scores used by KronRLS, (iii) DeepDTA \citep{ozturk2018deepdta}, a multi-layered CNN that learns features through raw protein sequences and SMILES strings. In order to compare these models with ChemBoost, we trained them from scratch on the KIBA and BDB data sets and Table 3 reports their performance on the corresponding test sets. Here ChemBoost refers to Model (9) in \ref{tab:bdb_kiba_merged_res} and describes ligands with SMILESVec and proteins with a combination of SW and ligand-centric representation.

SimBoost performed better than KronRLS in terms of all evaluation metrics on both data sets. This is an expected outcome given that SimBoost relies on network-based features as well as the features KronRLS utilized, namely the SW and PubChem similarity scores. ChemBoost, on the other hand, obtained either similar to or higher scores than SimBoost on both data sets. Although both ChemBoost and SimBoost depend on the XGBoost algorithm and utilize SW vectors for protein representation, ChemBoost incorporates chemical word vectors for both ligand and protein representations, indicating the effectiveness of the information they bring in. On the BDB data set, DeepDTA obtained a better MSE than ChemBoost, but a similar CI (significance level 95\%), whereas on KIBA, ChemBoost achieved a better MSE but worse CI than DeepDTA with the 99\% significance
level, indicating that their prediction performances are comparable.

Hence, we can suggest that ChemBoost achieves state-of-the-art level performance by exploiting distributed chemical word vectors and protein sequence information.

\subsection{ChemBoost can capture functional similarity of proteins
with low sequence similarity}
\label{subsec:seq_sim}
\begin{figure*}
    \centering
    \begin{tabular}{@{}c@{}}
        \includegraphics[width=.5\textwidth]{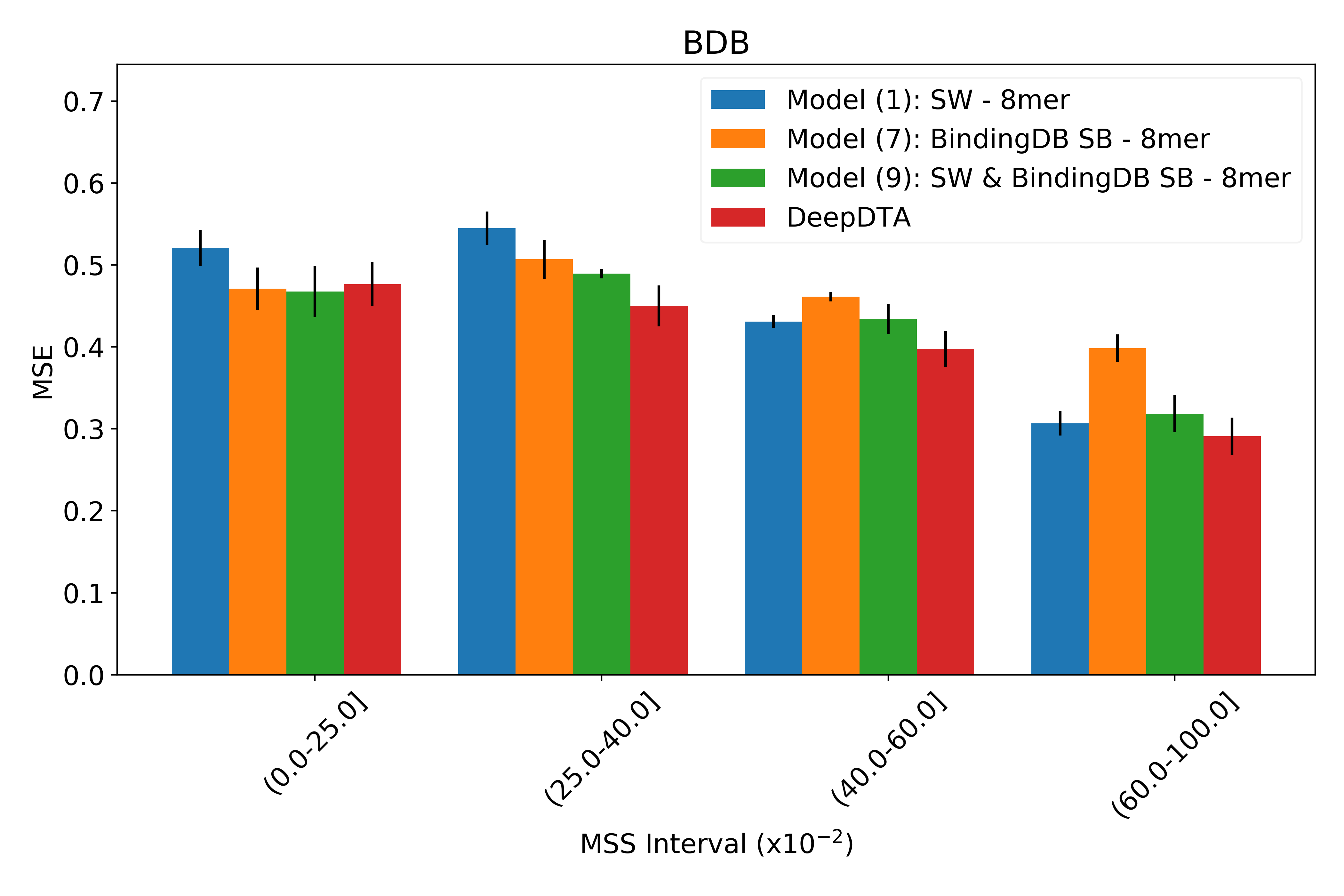} \includegraphics[width=.5\textwidth]{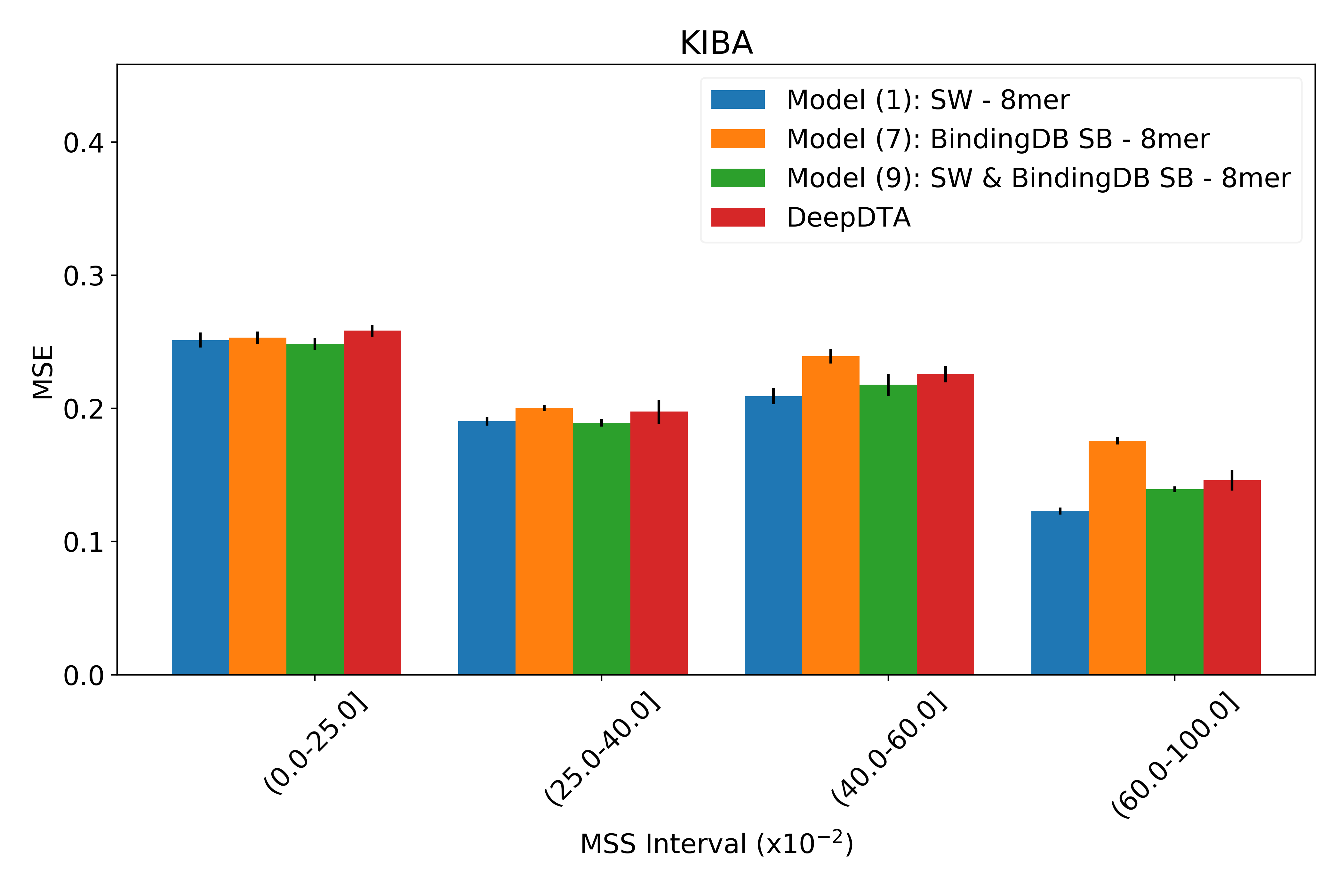}
    \end{tabular}
    \caption{Test set performance of ChemBoost and DeepDTA on BDB (left) and KIBA (right) with respect to $MSS$ of interactions.}
    \label{fig:seq_sim}
\end{figure*}

We investigated different ChemBoost models to observe the impact of using amino-acid sequence based protein representation in comparison to ligand-centric protein representation in predictive performance. Our focus was on proteins with low sequence similarity that bind to similar ligands. This case, where unrelated proteins bind to identical or similar ligands is complicated even in 3D space and no clear patterns emerge \citep{barelier2015recognition}. However, ligand binding is known to capture mechanistic information about the protein \citep{martin1998protein, hert2008quantifying} and a ligand centric approach is expected to boost performance in the protein - ligand interaction prediction task, especially for proteins of low sequence similarity.

Therefore, we investigated the performance of ChemBoost models as a function of protein sequence similarity. For each protein-ligand pair (P-L) in the test set, we computed the normalized S-W similarity score \citep{yamanishi2008predict} of P to the other interacting proteins of L in the training set. Then, we calculated the maximum score, which we refer to as Maximum Sequence Similarity ($MSS_{PL}$), for a P-L pair. We formulate $MSS_{PL}$ as:

$$MSS_{PL} = max\{SW(P,p) \forall p \in P(L) \}$$

\noindent where $P(L)$ the set of proteins with a reported affinity with
ligand L in the training set.

We divided the P-L pairs in the test set into 4 categories with respect to their $MSS$, such that each category comprises approximately the same number of interactions (around 5000 P-L pairs for KIBA and 1250 for BDB). Here, an interaction pair P-L being in a high $MSS$ category, such as (60.00, 100], indicates that the models already observed an interaction of
ligand L with a protein that is sequence-wise similar to P. Likewise, we can consider low $MSS$ pairs in test set as pairs with no similar interactions in the training set. 

For each $MSS$ category, we computed MSE value with Model (1), Model (7), Model (9) of ChemBoost and DeepDTA. All ChemBoost models use SMILESVec to describe ligands. For protein representation, however, Model (1) utilizes SW vectors, Model (7) uses a strictly ligand-centric approach with BindingDB integration and Model (9) combines both. Last, DeepDTA uses raw protein and ligand sequences and shown as the best performing model among the existing benchmarks in Section \ref{sec:sota}. We repeated the computations 5 times for each model using different training fold configurations and Figure \ref{fig:seq_sim} illustrates the mean and standard deviation of MSE scores for the given $MSS$ intervals for BDB (left) and KIBA (right) data sets.

For BDB, Model (1) that uses only SW vectors performs significantly worse than all three models (paired t-test, significance=95\%) in the lowest $MSS$ category, proposing that SW vectors were insufficient to capture functional similarity when sequence similarity is low. The same statistical test did not suggest a distinction for KIBA. On the other hand, we observed a clear performance increase of Model (1) in the highest $MSS$ category for both data sets.

Unlike Model (1), Model (7) is unaware of protein sequence information since it depends on SMILESVecs based ligand-centric representations to describe proteins. Model (7) is on par with the rest of the models in the lowest $MSS$ category of both data sets and second lowest category of KIBA, although its overall performance was significantly worse than all. This highlights the merit of using ligand-centric representations when functional similarity cannot be captured from protein sequence. 

Model (9) combines SW with chemical language based ligand-centric representations and performs comparably to DeepDTA as shown in Section \ref{sec:sota}. Here, we observe that Model (9) achieves relatively consistent scores across all $MSS$ categories. For all categories except the highest $MSS$ category of KIBA, Model (9) is on par with (99\% significance) the best model of the category. On the other hand, DeepDTA is negatively affected by low sequence similarity and presents an unstable prediction performance with higher standard deviation. These results show that, Model (9) can exploit the advantages of both SW and ligand-centric protein representations and is more consistent than DeepDTA.

In conclusion, we showed that the information encoded through sequence similarity and ligand-centric approach emphasizes different characteristics in protein representation. While ligand-centric models are able to capture functional similarity without using protein sequence information, SW-based models can exploit high sequence similarity. Combining these two complementary approaches, ChemBoost achieves state-of-the-art performance with robustness to the changes in protein sequence similarity.

\subsection{Evaluating ChemBoost on Novel Biomolecules}
In order to estimate the performance of ChemBoost for novel proteins and ligands, we performed experiments using different train-test splits. We randomly grouped the proteins and ligands in BDB and KIBA into two categories as known and unknown in order to create one training split and four test splits, per dataset. For BDB and KIBA, respectively, we divided the interactions of the molecules in the known group into two sets and obtained the training set and the first test set (warm). Afterwards, we identified the interactions of the known proteins with unknown ligands (cold ligand), unknown proteins with known ligands (cold protein), and unknown proteins with unknown ligands to form the second, third, and fourth test sets, respectively \citep{feng2018padme}. We respected the size of the training sets to be close the ones in our previous experimental setup and size of each test set to be close to each other.

We trained Model (1), Model (7), and Model (9), on the new training set and tested them on each test set. Here, the cold-protein representation is especially challenging for Model (7) and Model (9), since they utilize the high-affinity ligands of proteins for protein representation, but the proteins are absent in the training set. We alleviate this problem by using the ligand-centric vector of the sequence-wise most similar (based on 3-mer based Jaccard similarity) protein in the training set to compute ligand-centric vectors for cold proteins. We also evaluated the performance of DeepDTA as a strong benchmark using the same train-test configuration. We repeated the random splitting and model training 5 times and report the results in Table 4, for both data sets.

For the warm test set, we observed that the ranking of the models is similar to our previous results, validating that the new experimental setup is also reliable. On the other hand, the results for cold ligand split did not indicate an apparent ordering of the models, suggesting that SMILESVec representations, learned from an external corpus, are as good as the CNN-based representations learned from the training sets to represent novel ligands. We then analyzed the results on the cold protein split and observed that Model (1) and Model (9), which utilize SW vectors for protein representation, obtained higher results than Model (7). This shows that SW vectors are more robust to represent novel proteins, compared to ligand-centric representations. We then analyzed the scores on the cold split and observed that all models performed worse than the first three test sets. This indicates that the interactions where the protein and ligand are both novel, are challenging for all affinity prediction models.

\begin{table}
\caption{Performance of three ChemBoost models and DeepDTA on the interactions of known and unknown molecules in BDB (top) and KIBA (bottom) data sets. We report mean MSE and CI scores on each test split based on 5 different random splits of the molecules. Standard deviations are reported in parentheses and the naming of the test sets is similar to the one used by \citealp{feng2018padme}}
\begin{center}
\resizebox{\textwidth}{!}{

\begin{tabular}{llcccccccc}

\toprule
& & \multicolumn{2}{c} {\textbf { Warm }} & \multicolumn{2}{c} {\textbf { Cold Ligand }} & \multicolumn{2}{c} {\textbf { Cold Protein }} & \multicolumn{2}{c} {\textbf { Cold }} \\

& \textbf{ Model } & \textbf{ MSE } & \textbf{ CI } & \textbf{ MSE } & \textbf{ CI } & \textbf{ MSE } & \textbf{ CI } & \textbf{ MSE } & \textbf{ CI } \\ \midrule

\parbox[t]{2mm}{\multirow{4}{*}{\rotatebox[origin=c]{90}{\textbf{BDB}}}} & Model (1) &  0.373 (0.016) &  0.885 (0.010)  & 1.178 (0.079) &  0.736 (0.036) &  0.720 (0.094) &  0.799 (0.012) &  1.393 (0.145)  & 0.657 (0.044) \\

& Model (7) &  0.404 (0.010) &  0.863 (0.010) &  1.185 (0.143) &  0.700 (0.044) &  1.156 (0.251) &  0.749 (0.023) &  1.576 (0.185) &  0.596 (0.055) \\

& Model (9) &  0.361 (0.010) &  0.880 (0.008) &  1.157 (0.285) &  0.730 (0.044) &  0.800 (0.145) &  0.786 (0.023) &  1.358 (0.324) &  0.665 (0.053) \\

& DeepDTA  & 0.345 (0.026) &  0.879 (0.007) &  1.350 (0.306) &  0.672 (0.024) &  0.810 (0.147) &  0.778 (0.015) &  1.522 (0.300) &  0.614 (0.039) \\ \midrule

\parbox[t]{2mm}{\multirow{4}{*}{\rotatebox[origin=c]{90}{\textbf{KIBA}}}} & Model (1) &  0.185 (0.010) &  0.845 (0.006) &  0.450 (0.040) &  0.732 (0.009) &  0.298 (0.024) &  0.762 (0.031) &  0.588 (0.058) &  0.646 (0.043) \\

 & Model (7) &  0.202 (0.008) &  0.839 (0.005) &  0.445 (0.038) &  0.736 (0.009) &  0.453 (0.051) &  0.734 (0.018) &  0.667 (0.060) &  0.638 (0.027) \\

 & Model (9) &  0.183 (0.006) &  0.847 (0.005) &  0.442 (0.034) &  0.735 (0.011) &  0.340 (0.011)  & 0.748 (0.021) &  0.614 (0.047) &  0.640 (0.033) \\

 & DeepDTA &  0.199 (0.014) &  0.853 (0.005) &  0.456 (0.068) &  0.754 (0.012) &  0.400 (0.054)  & 0.747 (0.020) &  0.655 (0.080) &  0.652 (0.045) \\
\bottomrule
\end{tabular}
}
\end{center}
\end{table}

\section{Conclusion} \label{sec:conclusion}

In this work, we introduce ChemBoost, a chemical-word embeddings based affinity prediction framework. ChemBoost achieves state-of-the-art prediction performance, when ligands are represented with SMILESVecs and proteins with the combination of SW and the chemical words of their high affinity ligands. We evaluated the ChemBoost models on the BDB and KIBA data sets using MSE and CI.

As chemical words are at the core of our framework, we compared 8-mers with BPE words. The experiments showed that 8-mers create better ligand and protein representations for affinity prediction and we recommend their usage as they are also simpler.

Rost proposed the presence of a ``twilight zone" in sequence similarity, for which information about the protein can be predicted only in the presence of additional information \citep{rost}. We show that, SW is not able to accurately capture binding information when sequence similarity is low and a ligand-centric approach can improve model performance, especially for proteins with low sequence similarity. When used as a standalone representation, ligand-centric protein representation is more successful than SW at capturing functional similarities for low $MSS$ interactions. In representing novel biomolecules though, the combination of SW and chemical language based ligand vectors are more successful than the representations learned from the training set. We also observed that ligand-centric representations are more powerful when only high affinity ligands are used and when the number of high affinity ligands per protein is increased.
The ligand centric approach is limited to proteins with ligand binding information and it can be used in combination with orthogonal pieces of information for tasks ranging from fold prediction \citep{khor2015general} to function annotation \citep{gana2013structural}. Model (9), which uses both SW and ligand-centric vectors achieved state-of-the-art performance and is more robust to the changes in sequence similarity than both Model (1) and the current state-of-the-art model, DeepDTA.

Our results showed that good ligand and protein representations are essential for success in binding affinity prediction. Our approach, built on chemical language based representations of compounds and their interacting proteins, can be modified to use different chemical words, different compound representations or even features that utilize domain expertise.

\section*{Acknowledgement}
This work is partially supported by TUBITAK (The Scientic and Technological
Research Council of Turkey) under grant number 119E133. TUBITAK-BIDEB 2211-A Scholarship Program (to R.O.), TUBITAK-BIDEB 2211-E Scholarship Program (to H.O.) and TUBA-GEBIP Award of the Turkish Science Academy (to A.O.) are gratefully acknowledged. E.O. thanks Prof Amedeo Caflisch for hosting her at University of Zurich during her sabbatical. Authors thank Selen Parlar for her help with the calculations.

\bibliography{references}

\begin{thebibliography}{55}
\providecommand{\natexlab}[1]{#1}
\providecommand{\url}[1]{\texttt{#1}}
\expandafter\ifx\csname urlstyle\endcsname\relax
  \providecommand{\doi}[1]{doi: #1}\else
  \providecommand{\doi}{doi: \begingroup \urlstyle{rm}\Url}\fi

\bibitem[Apweiler et~al.(2004)Apweiler, Bairoch, Wu, Barker, Boeckmann, Ferro,
  Gasteiger, Huang, Lopez, Magrane, et~al.]{apweiler2004uniprot}
R.~Apweiler, A.~Bairoch, C.~H. Wu, W.~C. Barker, B.~Boeckmann, S.~Ferro,
  E.~Gasteiger, H.~Huang, R.~Lopez, M.~Magrane, et~al.
\newblock Uniprot: the universal protein knowledgebase.
\newblock \emph{Nucleic acids research}, 32\penalty0 (suppl\_1):\penalty0
  D115--D119, 2004.

\bibitem[Asgari and Mofrad(2015)]{asgari2015continuous}
E.~Asgari and M.~R. Mofrad.
\newblock Continuous distributed representation of biological sequences for
  deep proteomics and genomics.
\newblock \emph{PloS one}, 10\penalty0 (11):\penalty0 e0141287, 2015.

\bibitem[Barcellos et~al.(2019)Barcellos, Santos, Federico, Almeida, da~Silva,
  and Taft]{barcellos2019pharmacophore}
M.~P. Barcellos, C.~B. Santos, L.~B. Federico, P.~F.~d. Almeida, C.~H. d.~P.
  da~Silva, and C.~A. Taft.
\newblock Pharmacophore and structure-based drug design, molecular dynamics and
  admet/tox studies to design novel potential pad4 inhibitors.
\newblock \emph{Journal of Biomolecular Structure and Dynamics}, 37\penalty0
  (4):\penalty0 966--981, 2019.

\bibitem[Barelier et~al.(2015)Barelier, Sterling, O’Meara, and
  Shoichet]{barelier2015recognition}
S.~Barelier, T.~Sterling, M.~J. O’Meara, and B.~K. Shoichet.
\newblock The recognition of identical ligands by unrelated proteins.
\newblock \emph{ACS chemical biology}, 10\penalty0 (12):\penalty0 2772--2784,
  2015.

\bibitem[Barley et~al.(2018)Barley, Turner, and Goodacre]{barley2018improved}
M.~H. Barley, N.~J. Turner, and R.~Goodacre.
\newblock Improved descriptors for the quantitative structure--activity
  relationship modeling of peptides and proteins.
\newblock \emph{Journal of chemical information and modeling}, 58\penalty0
  (2):\penalty0 234--243, 2018.

\bibitem[Bosc et~al.(2019)Bosc, Atkinson, Felix, Gaulton, Hersey, and
  Leach]{bosc2019large}
N.~Bosc, F.~Atkinson, E.~Felix, A.~Gaulton, A.~Hersey, and A.~R. Leach.
\newblock Large scale comparison of qsar and conformal prediction methods and
  their applications in drug discovery.
\newblock \emph{Journal of cheminformatics}, 11\penalty0 (1):\penalty0 4, 2019.

\bibitem[Bostr{\"o}m et~al.(2018)Bostr{\"o}m, Brown, Young, and
  Keser{\"u}]{bostrom2018expanding}
J.~Bostr{\"o}m, D.~G. Brown, R.~J. Young, and G.~M. Keser{\"u}.
\newblock Expanding the medicinal chemistry synthetic toolbox.
\newblock \emph{Nature Reviews Drug Discovery}, 2018.

\bibitem[Cadeddu et~al.(2014)Cadeddu, Wylie, Jurczak, Wampler-Doty, and
  Grzybowski]{cadeddu2014organic}
A.~Cadeddu, E.~K. Wylie, J.~Jurczak, M.~Wampler-Doty, and B.~A. Grzybowski.
\newblock Organic chemistry as a language and the implications of chemical
  linguistics for structural and retrosynthetic analyses.
\newblock \emph{Angewandte Chemie International Edition}, 53\penalty0
  (31):\penalty0 8108--8112, 2014.

\bibitem[Chen and Guestrin(2016)]{chen2016xgboost}
T.~Chen and C.~Guestrin.
\newblock Xgboost: A scalable tree boosting system.
\newblock In \emph{Proceedings of the 22nd acm sigkdd international conference
  on knowledge discovery and data mining}, pages 785--794. ACM, 2016.

\bibitem[Chen et~al.(2018)Chen, Zhao, Li, Leier, Marquez-Lago, Wang, Webb,
  Smith, Daly, Chou, et~al.]{chen2018ifeature}
Z.~Chen, P.~Zhao, F.~Li, A.~Leier, T.~T. Marquez-Lago, Y.~Wang, G.~I. Webb,
  A.~I. Smith, R.~J. Daly, K.-C. Chou, et~al.
\newblock ifeature: a python package and web server for features extraction and
  selection from protein and peptide sequences.
\newblock \emph{Bioinformatics}, 34\penalty0 (14):\penalty0 2499--2502, 2018.

\bibitem[Convard et~al.(1994)Convard, Dubost, Le~Solleu, and
  Kummer]{convard1994smilogp}
T.~Convard, J.-P. Dubost, H.~Le~Solleu, and E.~Kummer.
\newblock Smilogp: a program for a fast evaluation of theoretical log p from
  the smiles code of a molecule.
\newblock \emph{Quantitative Structure-Activity Relationships}, 13\penalty0
  (1):\penalty0 34--37, 1994.

\bibitem[Davies et~al.(2015)Davies, Nowotka, Papadatos, Dedman, Gaulton,
  Atkinson, Bellis, and Overington]{davies2015chembl}
M.~Davies, M.~Nowotka, G.~Papadatos, N.~Dedman, A.~Gaulton, F.~Atkinson,
  L.~Bellis, and J.~P. Overington.
\newblock Chembl web services: streamlining access to drug discovery data and
  utilities.
\newblock \emph{Nucleic acids research}, 43\penalty0 (W1):\penalty0 W612--W620,
  2015.

\bibitem[Durant et~al.(2002)Durant, Leland, Henry, and
  Nourse]{durant2002reoptimization}
J.~L. Durant, B.~A. Leland, D.~R. Henry, and J.~G. Nourse.
\newblock Reoptimization of mdl keys for use in drug discovery.
\newblock \emph{Journal of chemical information and computer sciences},
  42\penalty0 (6):\penalty0 1273--1280, 2002.

\bibitem[Feng et~al.(2018)Feng, Dueva, Cherkasov, and Ester]{feng2018padme}
Q.~Feng, E.~Dueva, A.~Cherkasov, and M.~Ester.
\newblock Padme: A deep learning-based framework for drug-target interaction
  prediction.
\newblock \emph{arXiv preprint arXiv:1807.09741}, 2018.

\bibitem[Friedman(2001)]{friedman2001greedy}
J.~H. Friedman.
\newblock Greedy function approximation: a gradient boosting machine.
\newblock \emph{Annals of statistics}, pages 1189--1232, 2001.

\bibitem[Gage(1994)]{gage1994new}
P.~Gage.
\newblock A new algorithm for data compression.
\newblock \emph{The C Users Journal}, 12\penalty0 (2):\penalty0 23--38, 1994.

\bibitem[Gana et~al.(2013)Gana, Rao, Huang, Wu, and
  Vasudevan]{gana2013structural}
R.~Gana, S.~Rao, H.~Huang, C.~Wu, and S.~Vasudevan.
\newblock Structural and functional studies of s-adenosyl-l-methionine binding
  proteins: a ligand-centric approach.
\newblock \emph{BMC structural biology}, 13\penalty0 (1):\penalty0 6, 2013.

\bibitem[Garfield(1961)]{garfield1961chemico}
E.~Garfield.
\newblock Chemico-linguistics: computer translation of chemical nomenclature.
\newblock \emph{Nature}, 192\penalty0 (4798):\penalty0 192, 1961.

\bibitem[Gomes et~al.(2017)Gomes, Ramsundar, Feinberg, and
  Pande]{gomes2017atomic}
J.~Gomes, B.~Ramsundar, E.~N. Feinberg, and V.~S. Pande.
\newblock Atomic convolutional networks for predicting protein-ligand binding
  affinity.
\newblock \emph{arXiv preprint arXiv:1703.10603}, 2017.

\bibitem[G{\"o}nen and Heller(2005)]{gonen2005concordance}
M.~G{\"o}nen and G.~Heller.
\newblock Concordance probability and discriminatory power in proportional
  hazards regression.
\newblock \emph{Biometrika}, 92\penalty0 (4):\penalty0 965--970, 2005.

\bibitem[He et~al.(2017)He, Heidemeyer, Ban, Cherkasov, and
  Ester]{he2017simboost}
T.~He, M.~Heidemeyer, F.~Ban, A.~Cherkasov, and M.~Ester.
\newblock Simboost: a read-across approach for predicting drug--target binding
  affinities using gradient boosting machines.
\newblock \emph{Journal of cheminformatics}, 9\penalty0 (1):\penalty0 24, 2017.

\bibitem[Hert et~al.(2008)Hert, Keiser, Irwin, Oprea, and
  Shoichet]{hert2008quantifying}
J.~Hert, M.~J. Keiser, J.~J. Irwin, T.~I. Oprea, and B.~K. Shoichet.
\newblock Quantifying the relationships among drug classes.
\newblock \emph{Journal of chemical information and modeling}, 48\penalty0
  (4):\penalty0 755--765, 2008.

\bibitem[Jaeger et~al.(2017)Jaeger, Fulle, and Turk]{jaeger2017mol2vec}
S.~Jaeger, S.~Fulle, and S.~Turk.
\newblock Mol2vec: Unsupervised machine learning approach with chemical
  intuition.
\newblock \emph{Journal of chemical information and modeling}, 2017.

\bibitem[Jim{\'e}nez~Luna et~al.(2018)Jim{\'e}nez~Luna, Skalic,
  Martinez-Rosell, and De~Fabritiis]{jimenez2018k}
J.~Jim{\'e}nez~Luna, M.~Skalic, G.~Martinez-Rosell, and G.~De~Fabritiis.
\newblock K deep: Protein-ligand absolute binding affinity prediction via
  3d-convolutional neural networks.
\newblock \emph{Journal of chemical information and modeling}, 2018.

\bibitem[Jones et~al.(2001)Jones, Oliphant, Peterson, et~al.]{jones2001scipy}
E.~Jones, T.~Oliphant, P.~Peterson, et~al.
\newblock Scipy: Open source scientific tools for python.
\newblock 2001.

\bibitem[Keiser et~al.(2007)Keiser, Roth, Armbruster, Ernsberger, Irwin, and
  Shoichet]{keiser2007relating}
M.~J. Keiser, B.~L. Roth, B.~N. Armbruster, P.~Ernsberger, J.~J. Irwin, and
  B.~K. Shoichet.
\newblock Relating protein pharmacology by ligand chemistry.
\newblock \emph{Nature biotechnology}, 25\penalty0 (2):\penalty0 197, 2007.

\bibitem[Keiser et~al.(2009)Keiser, Setola, Irwin, Laggner, Abbas, Hufeisen,
  Jensen, Kuijer, Matos, Tran, et~al.]{keiser2009}
M.~J. Keiser, V.~Setola, J.~J. Irwin, C.~Laggner, A.~I. Abbas, S.~J. Hufeisen,
  N.~H. Jensen, M.~B. Kuijer, R.~C. Matos, T.~B. Tran, et~al.
\newblock Predicting new molecular targets for known drugs.
\newblock \emph{Nature}, 462\penalty0 (7270):\penalty0 175--181, 2009.

\bibitem[Khor et~al.(2015)Khor, Tye, Lim, and Choong]{khor2015general}
B.~Y. Khor, G.~J. Tye, T.~S. Lim, and Y.~S. Choong.
\newblock General overview on structure prediction of twilight-zone proteins.
\newblock \emph{Theoretical Biology and Medical Modelling}, 12\penalty0
  (1):\penalty0 15, 2015.

\bibitem[Krallinger et~al.(2017)Krallinger, Rabal, Lourenco, Oyarzabal, and
  Valencia]{krallinger2017information}
M.~Krallinger, O.~Rabal, A.~Lourenco, J.~Oyarzabal, and A.~Valencia.
\newblock Information retrieval and text mining technologies for chemistry.
\newblock \emph{Chemical reviews}, 117\penalty0 (12):\penalty0 7673--7761,
  2017.

\bibitem[Landrum et~al.(2006)]{landrum2006rdkit}
G.~Landrum et~al.
\newblock Rdkit: Open-source cheminformatics, 2006.

\bibitem[Limongelli(2020)]{limongelliligand}
V.~Limongelli.
\newblock Ligand binding free energy and kinetics calculation in 2020.
\newblock \emph{Wiley Interdisciplinary Reviews: Computational Molecular
  Science}, page e1455, 2020.

\bibitem[Martin et~al.(1998)Martin, Orengo, Hutchinson, Jones, Karmirantzou,
  Laskowski, Mitchell, Taroni, and Thornton]{martin1998protein}
A.~C. Martin, C.~A. Orengo, E.~G. Hutchinson, S.~Jones, M.~Karmirantzou, R.~A.
  Laskowski, J.~B. Mitchell, C.~Taroni, and J.~M. Thornton.
\newblock Protein folds and functions.
\newblock \emph{Structure}, 6\penalty0 (7):\penalty0 875--884, 1998.

\bibitem[Mayr et~al.(2018)Mayr, Klambauer, Unterthiner, Steijaert, Wegner,
  Ceulemans, Clevert, and Hochreiter]{mayr2018large}
A.~Mayr, G.~Klambauer, T.~Unterthiner, M.~Steijaert, J.~K. Wegner,
  H.~Ceulemans, D.-A. Clevert, and S.~Hochreiter.
\newblock Large-scale comparison of machine learning methods for drug target
  prediction on chembl.
\newblock \emph{Chemical Science}, 2018.

\bibitem[Mikolov et~al.(2013)Mikolov, Sutskever, Chen, Corrado, and
  Dean]{mikolov2013distributed}
T.~Mikolov, I.~Sutskever, K.~Chen, G.~S. Corrado, and J.~Dean.
\newblock Distributed representations of words and phrases and their
  compositionality.
\newblock In \emph{Advances in neural information processing systems}, pages
  3111--3119, 2013.

\bibitem[Morgan(1965)]{morgan1965generation}
H.~L. Morgan.
\newblock The generation of a unique machine description for chemical
  structures-a technique developed at chemical abstracts service.
\newblock \emph{Journal of Chemical Documentation}, 5\penalty0 (2):\penalty0
  107--113, 1965.

\bibitem[{\"O}zt{\"u}rk et~al.(2015){\"O}zt{\"u}rk, Ozkirimli, and
  {\"O}zg{\"u}r]{ozturk2015classification}
H.~{\"O}zt{\"u}rk, E.~Ozkirimli, and A.~{\"O}zg{\"u}r.
\newblock Classification of beta-lactamases and penicillin binding proteins
  using ligand-centric network models.
\newblock \emph{PloS one}, 10\penalty0 (2):\penalty0 e0117874, 2015.

\bibitem[{\"O}zt{\"u}rk et~al.(2018{\natexlab{a}}){\"O}zt{\"u}rk,
  {\"O}zg{\"u}r, and Ozkirimli]{ozturk2018deepdta}
H.~{\"O}zt{\"u}rk, A.~{\"O}zg{\"u}r, and E.~Ozkirimli.
\newblock Deepdta: deep drug--target binding affinity prediction.
\newblock \emph{Bioinformatics}, 34\penalty0 (17):\penalty0 i821--i829,
  2018{\natexlab{a}}.

\bibitem[{\"O}zt{\"u}rk et~al.(2018{\natexlab{b}}){\"O}zt{\"u}rk, Ozkirimli,
  and {\"O}zg{\"u}r]{ozturk2018novel}
H.~{\"O}zt{\"u}rk, E.~Ozkirimli, and A.~{\"O}zg{\"u}r.
\newblock A novel methodology on distributed representations of proteins using
  their interacting ligands.
\newblock \emph{Bioinformatics}, 34\penalty0 (13):\penalty0 i295--i303,
  2018{\natexlab{b}}.
\newblock \doi{10.1093/bioinformatics/bty287}.

\bibitem[{\"O}zt{\"u}rk et~al.(2020){\"O}zt{\"u}rk, {\"O}zg{\"u}r, Schwaller,
  Laino, and Ozkirimli]{ozturk2020review}
H.~{\"O}zt{\"u}rk, A.~{\"O}zg{\"u}r, P.~Schwaller, T.~Laino, and E.~Ozkirimli.
\newblock Exploring chemical space using natural language processing
  methodologies for drug discovery.
\newblock \emph{Accepted for Drug Discovery Today}, 2020.

\bibitem[Pahikkala et~al.(2014)Pahikkala, Airola, Pietil{\"a}, Shakyawar,
  Szwajda, Tang, and Aittokallio]{pahikkala2014toward}
T.~Pahikkala, A.~Airola, S.~Pietil{\"a}, S.~Shakyawar, A.~Szwajda, J.~Tang, and
  T.~Aittokallio.
\newblock Toward more realistic drug--target interaction predictions.
\newblock \emph{Briefings in bioinformatics}, page bbu010, 2014.

\bibitem[Rehurek and Sojka(2011)]{rehurek2011gensim}
R.~Rehurek and P.~Sojka.
\newblock Gensim--python framework for vector space modelling.
\newblock \emph{NLP Centre, Faculty of Informatics, Masaryk University, Brno,
  Czech Republic}, 3\penalty0 (2), 2011.

\bibitem[Rost(1999)]{rost}
B.~Rost.
\newblock {Twilight zone of protein sequence alignments}.
\newblock \emph{Protein Engineering, Design and Selection}, 12\penalty0
  (2):\penalty0 85--94, 02 1999.
\newblock ISSN 1741-0126.
\newblock \doi{10.1093/protein/12.2.85}.
\newblock URL \url{https://doi.org/10.1093/protein/12.2.85}.

\bibitem[Schwaller et~al.(2018)Schwaller, Gaudin, Lanyi, Bekas, and
  Laino]{schwaller2018found}
P.~Schwaller, T.~Gaudin, D.~Lanyi, C.~Bekas, and T.~Laino.
\newblock “found in translation”: predicting outcomes of complex organic
  chemistry reactions using neural sequence-to-sequence models.
\newblock \emph{Chemical science}, 9\penalty0 (28):\penalty0 6091--6098, 2018.

\bibitem[Sennrich et~al.(2015)Sennrich, Haddow, and Birch]{sennrich2015neural}
R.~Sennrich, B.~Haddow, and A.~Birch.
\newblock Neural machine translation of rare words with subword units.
\newblock \emph{arXiv preprint arXiv:1508.07909}, 2015.

\bibitem[Sheridan et~al.(2016)Sheridan, Wang, Liaw, Ma, and
  Gifford]{sheridan2016extreme}
R.~P. Sheridan, W.~M. Wang, A.~Liaw, J.~Ma, and E.~M. Gifford.
\newblock Extreme gradient boosting as a method for quantitative
  structure--activity relationships.
\newblock \emph{Journal of chemical information and modeling}, 56\penalty0
  (12):\penalty0 2353--2360, 2016.

\bibitem[{\'S}led{\'z} and Caflisch(2018)]{sledz2018protein}
P.~{\'S}led{\'z} and A.~Caflisch.
\newblock Protein structure-based drug design: from docking to molecular
  dynamics.
\newblock \emph{Current opinion in structural biology}, 48:\penalty0 93--102,
  2018.

\bibitem[Stepniewska-Dziubinska et~al.(2018)Stepniewska-Dziubinska,
  Zielenkiewicz, and Siedlecki]{stepniewska2018development}
M.~M. Stepniewska-Dziubinska, P.~Zielenkiewicz, and P.~Siedlecki.
\newblock Development and evaluation of a deep learning model for
  protein-ligand binding affinity prediction.
\newblock \emph{Bioinformatics}, 1:\penalty0 9, 2018.

\bibitem[Tang et~al.(2014)Tang, Szwajda, Shakyawar, Xu, Hintsanen, Wennerberg,
  and Aittokallio]{tang2014making}
J.~Tang, A.~Szwajda, S.~Shakyawar, T.~Xu, P.~Hintsanen, K.~Wennerberg, and
  T.~Aittokallio.
\newblock Making sense of large-scale kinase inhibitor bioactivity data sets: a
  comparative and integrative analysis.
\newblock \emph{Journal of Chemical Information and Modeling}, 54\penalty0
  (3):\penalty0 735--743, 2014.

\bibitem[Wang et~al.(2004)Wang, Fang, Lu, and Wang]{wang2004pdbbind}
R.~Wang, X.~Fang, Y.~Lu, and S.~Wang.
\newblock The pdbbind database: Collection of binding affinities for protein-
  ligand complexes with known three-dimensional structures.
\newblock \emph{Journal of medicinal chemistry}, 47\penalty0 (12):\penalty0
  2977--2980, 2004.

\bibitem[Wang et~al.(2019)Wang, You, Yang, Li, Jiang, and Zhou]{wang2019high}
Y.~Wang, Z.-H. You, S.~Yang, X.~Li, T.-H. Jiang, and X.~Zhou.
\newblock A high efficient biological language model for predicting
  protein--protein interactions.
\newblock \emph{Cells}, 8\penalty0 (2):\penalty0 122, 2019.

\bibitem[Wo{\'z}niak et~al.(2018)Wo{\'z}niak, Wo{\l}os, Modrzyk, G{\'o}rski,
  Winkowski, Bajczyk, Szymku{\'c}, Grzybowski, and Eder]{wozniak2018linguistic}
M.~Wo{\'z}niak, A.~Wo{\l}os, U.~Modrzyk, R.~L. G{\'o}rski, J.~Winkowski,
  M.~Bajczyk, S.~Szymku{\'c}, B.~A. Grzybowski, and M.~Eder.
\newblock Linguistic measures of chemical diversity and the “keywords” of
  molecular collections.
\newblock \emph{Scientific reports}, 8, 2018.

\bibitem[Wu et~al.(2018)Wu, Ramsundar, Feinberg, Gomes, Geniesse, Pappu,
  Leswing, and Pande]{wu2018moleculenet}
Z.~Wu, B.~Ramsundar, E.~N. Feinberg, J.~Gomes, C.~Geniesse, A.~S. Pappu,
  K.~Leswing, and V.~Pande.
\newblock Moleculenet: a benchmark for molecular machine learning.
\newblock \emph{Chemical science}, 9\penalty0 (2):\penalty0 513--530, 2018.

\bibitem[Xue et~al.(2019)Xue, Tian, Wang, Xia, and Wu]{xue2019discovery}
W.~Xue, J.~Tian, X.~S. Wang, J.~Xia, and S.~Wu.
\newblock Discovery of potent ptp1b inhibitors via structure-based drug design,
  synthesis and in vitro bioassay of norathyriol derivatives.
\newblock \emph{Bioorganic chemistry}, 86:\penalty0 224--234, 2019.

\bibitem[Yamanishi et~al.(2008)Yamanishi, Araki, Gutteridge, Honda, and
  Kanehisa]{yamanishi2008predict}
Y.~Yamanishi, M.~Araki, A.~Gutteridge, W.~Honda, and M.~Kanehisa.
\newblock Prediction of drug--target interaction networks from the integration
  of chemical and genomic spaces.
\newblock \emph{Bioinformatics}, 24\penalty0 (13):\penalty0 i232--i240, 2008.

\bibitem[Ying et~al.(2019)Ying, Bourgeois, You, Zitnik, and
  Leskovec]{gnne2019ying}
Z.~Ying, D.~Bourgeois, J.~You, M.~Zitnik, and J.~Leskovec.
\newblock Gnnexplainer: Generating explanations for graph neural networks.
\newblock In H.~Wallach, H.~Larochelle, A.~Beygelzimer, F.~d\textquotesingle
  Alch\'{e}-Buc, E.~Fox, and R.~Garnett, editors, \emph{Advances in Neural
  Information Processing Systems 32}, pages 9240--9251. Curran Associates,
  Inc., 2019.
\newblock URL
  \url{http://papers.nips.cc/paper/9123-gnnexplainer-generating-explanations-for-graph-neural-networks.pdf}.

\end{thebibliography}
\bibliographystyle{abbrvnat}
\end{document}